# Evolutionary Biclustering of Clickstream Data

**R.Rathipriya [1a] , Dr. K.Thangavel [1b], J.Bagyamani[2c]**
**[1] Department of Computer Science, Periyar University, Salem, Tamilnadu, India**
**[2]Department of Computer Science,Government Arts College, Dharmapuri, Tamilnadu, India**

### Abstract

Biclustering is a two way clustering approach involving simultaneous clustering along two dimensions of the data matrix. Finding biclusters of web objects (i.e. web users and web pages) is an emerging topic in the context of web usage mining. It overcomes the problem associated with traditional clustering methods by allowing automatic discovery of browsing pattern based on a subset of attributes. A coherent bicluster of clickstream data is a local browsing pattern such that users in bicluster exhibit correlated browsing pattern through a subset of pages of a web site. This paper proposed a new application of biclustering to web data using a combination of heuristics and meta-heuristics such as K-means, Greedy Search Procedure and Genetic Algorithms to identify the coherent browsing pattern. Experiment is conducted on the benchmark clickstream msnbc dataset from UCI repository. Results demonstrate the efficiency and beneficial outcome of the proposed method by correlating the users and pages of a web site in high degree. This approach shows excellent performance at finding high degree of overlapped coherent biclusters from web data.

.**Keywords:** — Biclustering, Clickstream data, Coherent Bicluster, Genetic Algorithm, Greedy Search Procedure, Web Mining.

## 1. Introduction

The World Wide Web is the one of the important media to store, share, and distribute information in the large scale. Nowadays web users are facing the problems of information overload and drowning due to the significant and rapid growth in the amount of information and the number of users. As a result, how to provide web users with more exactly needed information is becoming a critical issue in web-based information retrieval and web applications.

Web mining [5,15] discovers and extracts interesting pattern or knowledge from web data. It is classified into three types as web content mining, web structure, and web usage mining. Web usage mining is the intelligent data mining technique to mine clickstream data in order to extract usage patterns. These patterns are analyzed to determine user's behavior which is an important and challenging research area in the web usage mining.

Clickstream data is a sequence of Uniform Resource Locators (URLs) browsed by the user within a particular period of time. To discover pattern of group of users with similar interest and motivation for visiting the particular website can be found by clustering. Traditional clustering [6] is used to cluster the web users or web pages based on the existing similarities. When a clustering method is used for grouping users, it typically partitions users according to their similarity of browsing behavior under all pages. However, it is often the case that some users behave similarly only on a subset of pages and their behavior is not similar over the rest of the pages. Therefore, traditional clustering methods fail to identify such user groups.

To overcome this problem, concept of Biclustering or Coclustering was introduced. Biclustering[2-4] was first introduced by Hartigan and called it direct clustering[11]. The application of biclustering in web mining is ideal when users have multiple interest/behavior over different subsets of web pages. Biclustering attempts to cluster web user and web pages simultaneously based on the users' behavior recorded in the form of clickstream data. It identifies the subset of users which show similar interest/ behavior under a specific subset of web pages. These browsing pattern play vital role in E-commerce based applications. Recommender systems analyze patterns of user browsing interest and to provide personalized services which match user's interest in most business domains, benefiting both the user and the merchant.

The objective of the proposed method is to identify set of subgroup of users and set of subgroup of pages with maximum volume such that these users and pages are highly correlated.

The rest of the paper is organized as follows. Section 2 describes some of the biclustering approaches available in the literature. Methods and materials required for biclustering approach are described in the section 3. Section 4 focuses on the proposed Biclustering framework using Genetic Algorithm. Analysis of experimental results is discussed in the Section 5. Section 6 concludes the paper with features for future enhancements.

## 2. Related Work

Koutsonikola, V.A. et al.[13] proposed a bi-clustering approach for web data, which identifies groups of related web users and pages using spectral clustering method on both row and column dimensions. S. Araya et al.[14] proposed methodology for target group identification from web usage data which improved the customer relationship management e.g. financial services. Sujatha et al.[16] proposed a novel method to improve the cluster quality using Genetic Algorithm (GA) for web usage data. Guandong et al.[10] presented an algorithm using bipartite spectral clustering to extract bicluster from web users and pages and the impact of using various clustering algorithms is also investigated in that paper.

The followings are the some of the biclustering algorithm available in the literature. Cho et al. [4] introduced K-Means based biclustering algorithms that identifies ' m' row clusters and 'n' column clusters while montonically decreasing the Mean Square Residue score defined by Cheng and Church. Dhillon et al.[8] proposed an innovative co-clustering algorithm that monotonically increases the preserved mutual information by intertwining both the row and column clusterings at all stages. Tang et al.[17] introduced a framework for unsupervised analysis of gene expression data which applies an interrelated two-way clustering approach on the gene expression matrices. Kluger et al.[12] proposed a method to discover the biclusters with coherent values and looked for checkerboard structures in the data matrix by integrating biclustering of rows and columns with normalization of the data matrix. Another approach called Double Conjugated Clustering (DCC) which aims to discover biclusters with coherent values defined using multiplicative model of bicluster by Busygin et al.[3]

Coupled Two Way Clustering algorithm[9] was introduced by Getz et al. which performs one way clustering on the rows and columns of the data matrix using stable clusters of row as attributes for column clustering and vice versa. Bleuler et al. [2] propose a evolutionary algorithm framework that embeds a greedy strategy. Chakraborty et al. [5] use genetic algorithm to eliminate the threshold of the maximum allowable dissimilarity in a bicluster.

In literature, biclustering algorithms are widely applied to the gene expression data. Most of these algorithms are failed to extract the coherent pattern from the data matrix. In web mining, there is no related work that has been applied specific biclustering algorithms for discovering the coherent browsing patterns.

In this paper, Greedy Search Procedure and evolutionary approach namely Genetic Algorithm (GA) is introduced to obtain the optimal coherent browsing patterns. The results show that GA outperforms the greedy procedure by identifying coherent browsing patterns. These patterns are very useful in the decision making for target marketing.

## 3. Methods and Materials

### 3.1 Preprocessing

Clickstream data pattern is converted into web user access matrix A by using (1) in which rows represent users and columns represent pages of web sites. Let A( U, P) be an 'n x m' user access matrix where U be a set of users and P be a set of pages of a web site. It is used to describe the relationship between web pages and users who access these web pages. Let 'n' be the number of web user and 'm' be the number of web pages. The element $a_{ij}$ of A(U, P) represents frequency of the user $U_i$ of U visit the page $P_j$ of P during a given period of time.

$$a_{ij} = \begin{cases} \text{Hits}(U_i, P_j), & \text{if } P_j \text{ is visited by } U_i \\ 0, & \text{otherwise} \end{cases}$$

(1)

where $\text{Hits}(U_i, P_j)$ is the count/frequency of the user $U_i$ accesses the page $P_j$ during a given period of time.

### 3.2 Coherent Bicluster

A bicluster with coherent values is the subset of users and subsets of pages with coherent values on both rows and columns. A measure called Average Correlation Value (ACV)[1] is used to measure the degree of coherence of the biclusters.

### 3.3 Average Correlation Value

It is used to evaluate the homogeneity of a bicluster. Matrix B = $(b_{ij})$ has the ACV which is defined by the following function,

$$ACV(B) = \max\{ \frac{\sum_{i=1}^{n}\sum_{j=1}^{n}|r\_row_{ij}| - n}{n^2 - n}, \frac{\sum_{k=1}^{m}\sum_{l=1}^{m}|r\_col_{kl}| - m}{m^2 - m} \}$$

(2)

$r\_row_{ij}$ is the correlation between row i and row j,

$r\_col_{kl}$ and is the correlation between column k and column l. A high ACV suggests high similarities among the users or pages. ACV can tolerate translation as well as scaling. And also works well for biclusters in which there's a linear correlation among the users or pages.

### 3.4. Greedy Search Procedure

A greedy algorithm repeatedly executes a search procedure which tries to maximize the bicluster based on examining local conditions, with the hope that the outcome will lead to a desired outcome for the global problem. This approach employs simple strategies that are easy to implement and most of the time quite efficient.

**Structure of Greedy Search Procedure**
Step 1: Start with initial bicluster.
Step 2: For every iteration
 Add/ remove the element(user/page) to/from the bicluster which maximize  the objective function.
End for

In this paper, objective function is to maximize ACV of a bicluster.

### 3.5 Encoding of Biclusters

Each enlarged and refined bicluster is encoded as a binary string .The length of the string is the number of rows plus the number of columns of the user access matrix A (U, P). A bit is set to one when the corresponding user or page is included in the bicluster. These binary encoded biclusters are used as initial population for genetic algorithm.

### 3.6 Volume of Bicluster

The number of elements in bicluster B (I, J) is called the volume of bicluster B (I, J) and denoted as VOL (B (I, J)).

VOL (B (I, J)) = |I| × |J|                    (3)

where, |I| is the number of users in the B and |J| is the number of pages in B.

## 4.  Coherent Biclustering Approach Using Evolutionary Algorithm

The proposed algorithm is used to identify the optimal coherent biclusters in terms of volume and quality in three subsequent steps. First step is to identify the initial biclusters called seeds by using K-Means clustering algorithm. Second step is to enlarge and refine these seeds using greedy search procedure which results in local optimum. Third step is to obtain global optimum of biclusters using evolutionary technique called genetic algorithm. These overlapped coherent biclusters have high degree of correlation among subset of users and subset of related pages of a web site.

This algorithm identifies the coherent browsing pattern from the web usage data which plays vital role in the direct marketing and target marketing. One-to-one relation between web users and pages of a web site is not appropriate because web users are not strictly interested in one category of web pages. Therefore, the proposed algorithm is tuned to discover the overlapping coherent biclusters from clickstream data patterns.

### 4.1 Bicluster Formation using K-Means Algorithm

In this paper, K-Means clustering method is applied on the web user access matrix A(U, P) along both dimensions separately to generate $k_u$ user clusters and $k_p$ page clusters .And then combine the results to obtain small co-regulated submatrices ($k_u × k_p$) called biclusters. These correlated biclusters are called seeds.

### 4.2 Enlargement and Refinement of Bicluster Using Greedy Search Procedure

In this step, seeds are enlarged and refined by adding /removing the rows and columns to enlarge their volume and improve their quality respectively. The main goal of the greedy search procedure is to maximize the volume of the bicluster seed without degrading the quality measure.

Here, ACV is used as merit function to grow the seeds. Insert/Remove the users/pages to /from the bicluster if it increases ACV of the bicluster.

**Algorithm 1: Seed Enlargement and Refinement using Greedy Search Procedure**
   **Input:** User Access Matrix A
   **Output:** Set of enlarged and refined biclusters

Step 1.      Compute $k_u$ user clusters and $k_p$ page clusters from preprocessed clickstream data.
Step 2.      Combine $k_u$ and $k_p$ clusters to form $k_u ×$ $k_p$ biclusters called seeds.
Step 3.      For each seed do
              Call Seed Enlargement(Seed(U, P))

Call Seed Refinement(Seed( U, P))
Step 4.        Return enlarged and refined  biclusters

**Algorithm 2: Seed Enlargement (Seed(U, P))**
   **Input:** Set of seeds.
   **Output:** Set of enlarged seeds.
Step 1.        Set of users 'u' not in U
Step 2.        Set of pages 'p' not in P
Step 3.        For each node u/p do
             If ACV( union(Seed, u/p)) >
             ACV(Seed(U, P)) then
                 Add u/p to Seed(U, P)
             End(if)
         End(for)
Step 4.        Return Enlarged Seed

**Algorithm 3: Seed Refinement (Enlarged Seed(U, P))**
   **Input:** Set of seeds.
   **Output:** Set of refined seeds.

Step 1.        For each  node u/p in U/P
             Remove node  u/p from Enlarged Seed ,
             U'/P' be set of rows/columns in U/P but
             not contained u/p
             If  ACV (Enlarged Seed(U',  P')) >
             ACV(Enlarged Seed(U/p, P)
             Update U/P
     End(if)
 End(for)
     Step 2.        Return Refined seed as bicluster

Enlarging and refining the seed starts from page list followed by user list until ACV is increased using greedy search procedure.

## 4.3 Coherent Biclustering Framework using Genetic Algorithm (GA)

        The GA is a stochastic global search method that mimics the metaphor of natural biological evolution. GA operates on a population of potential solutions applying the principle of survival of the fittest to produce better and better approximations to a solution. At each generation, a new set of approximations is created by the process of selecting individuals according to their level of fitness in the problem domain and breeding them together using operators borrowed from natural genetics. This process leads to the evolution of populations of individuals that are better suited to their environment than the individuals that they were created from, just as in natural adaptation.

Biclustering approach is viewed as optimization problem with the objective of discovering overlapping coherent biclusters with high ACV and high volume. In this paper, Genetic Algorithm (GA) is used for optimization of bicluster. The important feature of GA is that it provides a number of potential solutions to a given problem and the choice of final solution is left to the user.

        Usually, GA is initialized with the population of random solutions. In order to avoid random interference, biclusters obtained from greedy search procedure are used to initialize GA. This will result in faster convergence compared to random initialization. Maintaining diversity in the population is another advantage of initializing with these biclusters.

Fitness Function
   The main objective of this work is to discover high volume biclusters with high ACV. The following fitness function F (I, J) is used  to extract optimal bicluster.

$$
F (I, J) = \begin{cases} |I|*|J| \text{ , if ACV(bicluster)} \geq \delta \\ \\ 0, \text{ Otherwise} \qquad\qquad (4) \end{cases}
$$

Where |I| and |J| are number of rows and columns of bicluster and  δ is defined as follows

   ACV threshold δ =  Max(ACV(P))

Here, the objective function should be maximized. P is the set of biclusters in each population, *mp* is the probability of mutation, *r* is the fraction of the population to be replaced by crossover in each population, *cp* is the fraction of the population to be replaced by crossover in each population, n is the number of biclusters in each population. The biclustering framework using genetic algorithm is given below.

**Algorithm 4: Evolutionary Biclustering Algorithm**
Input: Set enlarged and refined seed
Output: Optimal Bicluster
   Step 1.        Initialize the population
   Step 2.        Evaluate the fitness of individuals
   Step 3.        For *i* =1 to max_iteration
             Selection()
             Crossover()
             Mutation()
             Evaluate the fitness
         End(For)

Step 4.   Return the optimal bicluster

Selection: The most commonly used form of GA selection is Roulette Wheel Selection (RWS), is used for the selection operator. When using RWS, a certain number of biclusters of the next generation are selected probabilistically, where the probability of selecting a bicluster solution $S_{nn}$ is given by

$$Pr(S_n) = Fitness(S_n) \, / \sum\nolimits_{N=1...n} Fitness(S_n) \qquad (4)$$

With RWS, each solution will have a probability to survive by being assigned with a positive fitness value. A solution with a high volume has a greater fitness value and hence has a higher probability to survive. On the other side, weaker solutions also have a chance to survive the selection process. This is an advantage, as though a solution may be weak, it may still contain some useful components.

Crossover and Mutation: Then $cp$ of parents is chosen probabilistically from the current population and the crossover operator will produce two new offsprings for each pair of parents using one point crossover technique on genes and conditions separately. Now the new generation contains the desired number of members and the mutation will increase or decrease the membership degree of each user and page with a small probability of mutation $mp$.

# 5. Experimental Results and Analysis

The experiments are conducted on the well-known benchmark clickstream dataset called msnbc dataset which was collected from MSNBC.com portal. This data set is taken from UCI repository, where the original data is preprocessed using equation 1. There are 989,818 users and only 17 distinct items, because these items are recorded at the level of URL category, not at page level, which greatly reduces the dimensionality.

The length of the clickstream record starts from 1 to 64. Average number of visits per user is 5.7. Intuitively, very small and very large number of URL category visited may not provide any useful information about the user's behavior. Thus, the length of the record having less than 5 is considered as a very small and record length greater than 15 is considered as a very large. During data filtering process, small and large records are removed from the dataset.

The metric index R is used to evaluate the overlapping degree between biclusters. It quantifies the amount of overlapping among biclusters. Degree of overlapping[7], is used as quantitative index to evaluate quantitatively the quality of generated biclusters. The degree of overlapping among all biclusters is defined as follows

$$R = \frac{1}{|U|*|P|} \sum_{i=1}^{|U|} \sum_{j=1}^{|P|} T_{ij}$$

$where$

$$T_{ij} = \frac{1}{(N-1)} * \left( \sum_{k=1}^{N} W_k(a_{ij}) - 1 \right) \qquad (5)$$

where N is the total number of biclusters, |U| represents the total number of users, and |P| represents the total number of pages in the data matrix A. The value of $w_k(a_{ij})$ is either 0 or 1. If the element (point) $a_{ij}$   A is present in the $k^{th}$ bicluster, then $w_k(a_{ij}) = 1$, otherwise 0. Hence, the R index represents the degree of overlapping among the biclusters. If R index value is higher, then degree of overlapping of the generated biclusters would be high. The range of R index is $0 \le R \le 1$.

During the bicluster formation step, K-Means clustering algorithm is applied along the both dimensions to generate $k_u$ and $k_p$ clusters and combined these clusters to get $k_u* k_p$ initial biclusters called seeds. Seeds are the biclusters whose volume is small. During the second step, seeds are enlarged and refined iteratively using Greedy Search Procedure. These seeds are enlarged and refined at incremental of ACV to reach high volume which is evident from the Table 1 and Table 2.

Table 1: Performance of Biclustering using Greedy Search Procedure

|  | Seed Formation Phase | Seed Enlargement and Refinement Phase |
|---|---|---|
| No. of Seeds | 114 | 114 |
| Average ACV | 0.4711 | 0.9413 |
| Average Volume | 494.9 | 1599.8 |

Table 2: Step-Wise Performance of Biclustering using Greedy Search Procedure

|  | Average ACV | Average Volume |
|---|---|---|
| Initial Bicluster | 0.4711 | 494.9 |
| Seed After Column Insertion | 0.8420 | 758.2 |
| Seed After Row Insertion | 0.8848 | 3488.6 |
| Seed After Column deletion | 0.9395 | 1600.5 |
| Seed After Row deletion | 0.9413 | 1599.8 |

Each bicluster seed underwent four stages of seed enlargement and refinement step. During each stage their ACV is incremented which is shown in Fig 2. Since the quality of the bicluster is more important than the volume, the volume adjusted in order to achieve the high ACV in various stages of the second phase which is portraits in Fig1.

To avoid random interference, very tightly correlated biclusters obtained using greedy search procedures are used as initial population for GA. Moreover, it results in quick convergence and provides number of potential biclusters. These biclusters have high ACV and high volume which is obvious from table 4. This approach shows excellent performance at finding high degree of overlapped coherent biclusters from web data.

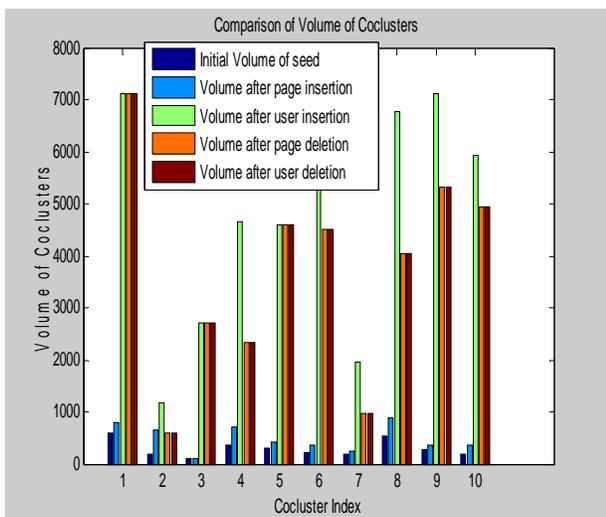

Fig 1. Volume of Biclusters in Various Stages

Table 3: Parameter Setting for GA

| Crossover Probability | 0.7 |
|---|---|
| Mutation Probability | 0.01 |
| Population Size | 114 |
| Generation | 100 |
| ACV Thersold | 0.95 |

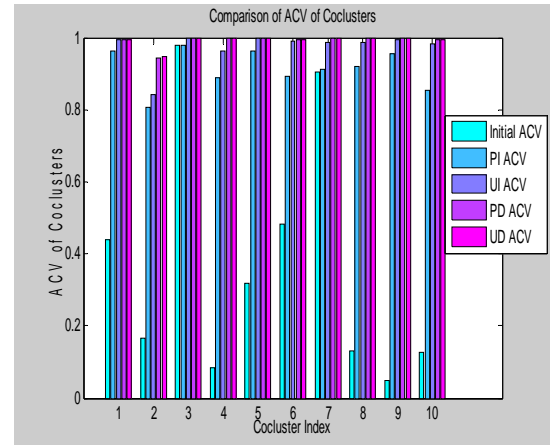

Fig 2. ACV of Biclusters in Various Stages

Table 4: Performance of Biclustering using GA

| Mean Volume | Mean ACV | Row Percent-age | Column percent-age | Overlapp-ing Degree |
|---|---|---|---|---|
| 12715 | 0.9609 | 99.9 | 82.35 | 0.2152 |

Table 5: Comparison of Average Volume and Homogeneity of biclusters

|  | Average Volume | Average ACV | Overlapping Degree |
|---|---|---|---|
| Two-Way K-Means | 494.9 | 0.4711 | 0 |
| Greedy Search Procedure | 1599.8 | 0.9413 | 0.0192 |
| Genetic Algorithm | 12715 | 0.9609 | 0.2152 |

These biclusters exhibit coherent pattern on a subset of dimensions. In clickstream analysis, the frequency of visiting the pages of a web site of two users may rise or fall synchronously in response to a set of their interest. Though the magnitude of their interest levels may not be close, but the pattern they exhibit can be very much similar. Our proposed biclustering frame work is interested in finding such coherent patterns of bicluster of users and with a general understanding of users' browsing interest. This method makes significant contribution in the field of web mining, E-Commerce applications and etc.

From the results, it is obvious that it correlates the relevant users and pages of a web site in high degree of homogeneity. Analyzing these overlapping coherent biclusters could be very beneficial for direct marketing, target marketing and also useful for recommending system, web personalization systems, web usage categorization and user profiling. The interpretation of biclustering results is also used by the company for focalized marketing campaigns to improve their performance of the business.

## CONCLUSION

The main contribution of this paper is twofold namely, development of coherent biclustering framework using GA to identify overlapped coherent biclusters from the clickstream data patterns and a coherence quality measure called ACV is used to get coherent biclusters in last two phases of the biclustering framework. The interpretation of the biclustering results can also be used towards improving the website's design, information availability and quality of provided services. The overlapping nature of the proposed framework can significantly contribute towards this direction. This method has potential to identify the coherent patterns automatically from the clickstream data. Future work aims at extending this framework by enriching clustering process would result to enhanced clusters' quality and a more accurate definition of relation coefficients.